\documentclass[10pt,twocolumn,letterpaper]{article}

\usepackage{iccv}
\usepackage{times}
\usepackage{epsfig}
\usepackage{graphicx}
\usepackage{amsmath}
\usepackage{amssymb}
\usepackage{multirow}

\usepackage{url}
\usepackage{times}
\usepackage{xcolor}
\usepackage{colortbl}
\usepackage{array} 
\usepackage{bm}
\usepackage{soul}
\usepackage{algorithm}
\usepackage{algorithmic}
\usepackage{multirow}
\usepackage{booktabs}
\usepackage{makecell}
\usepackage{bbding}
\usepackage{wrapfig}
\usepackage{bm}
\usepackage[accsupp]{axessibility} 

\usepackage{subcaption}
\usepackage[font=small]{caption} 

\usepackage[pagebackref=true,breaklinks=true,letterpaper=true,colorlinks,bookmarks=false]{hyperref}

\iccvfinalcopy 

\def\iccvPaperID{6768} 

\ificcvfinal\fi

\begin{document}

\title{MixReorg: Cross-Modal Mixed Patch Reorganization is a Good Mask Learner for Open-World Semantic Segmentation}

\author{Kaixin Cai$^{1}$\footnotemark[1] \quad Pengzhen Ren$^{1}$\footnotemark[1] \quad Yi Zhu$^{2}$ \quad Hang Xu$^{2}$ \quad Jianzhuang Liu$^{2}$ \\ 
Changlin Li$^{3}$ \quad Guangrun Wang$^{4}$ \quad Xiaodan Liang$^{1, 5, 6}$ \footnotemark[2] \\ 
{\normalsize $^{1}$Shenzhen Campus of Sun Yat-sen University  \quad $^{2}$Huawei Noah’s Ark Lab  \quad $^{3}$University of Technology Sydney} \\ 
{\normalsize $^{4}$University of Oxford  \quad $^{5}$MBZUAI  \quad  $^{6}$DarkMatter AI Research}\\ 
{\tt\small caikx7@mail2.sysu.edu.cn, pzhren@foxmail.com,  \{zhuyi36, xu.hang, liu.jianzhuang\}@huawei.com,}\\ 
{\tt\small \{changlinli.ai, wanggrun, xdliang328\}@gmail.com}}

\maketitle
\renewcommand{\thefootnote}{\fnsymbol{footnote}}
\footnotetext[1]{Equal contribution.}
\footnotetext[2]{Corresponding author.}

\ificcvfinal\thispagestyle{empty}\fi

\begin{abstract}
Recently, semantic segmentation models trained with image-level text supervision have shown promising results in challenging open-world scenarios. However, these models still face difficulties in learning fine-grained semantic alignment at the pixel level and predicting accurate object masks. To address this issue, we propose \textbf{MixReorg}, a novel and straightforward pre-training paradigm for semantic segmentation that enhances a model's ability to reorganize patches mixed across images, exploring both local visual relevance and global semantic coherence.
Our approach involves generating fine-grained patch-text pairs data by mixing image patches while preserving the correspondence between patches and text. The model is then trained to minimize the segmentation loss of the mixed images and the two contrastive losses of the original and restored features. With MixReorg as a mask learner, conventional text-supervised semantic segmentation models can achieve highly generalizable pixel-semantic alignment ability, which is crucial for open-world segmentation.
After training with large-scale image-text data, MixReorg models can be applied directly to segment visual objects of arbitrary categories, without the need for further fine-tuning. Our proposed framework demonstrates strong performance on popular zero-shot semantic segmentation benchmarks, outperforming GroupViT by significant margins of 5.0\%, 6.2\%, 2.5\%, and 3.4\% mIoU on PASCAL VOC2012, PASCAL Context, MS COCO, and ADE20K, respectively.
\end{abstract}

\begin{figure}[t]
    \centering
    \includegraphics[width=1 \linewidth]{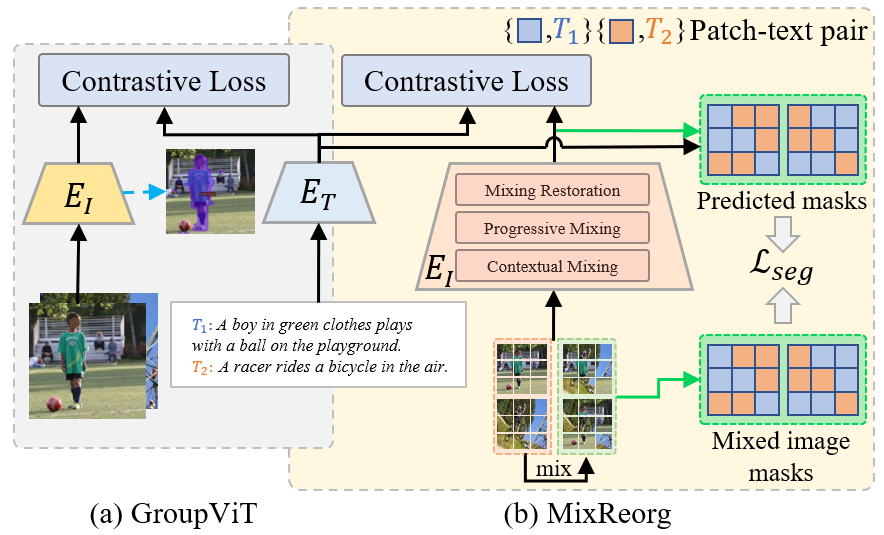}
    \vspace{-2em}
    \caption{
    Comparison between GroupViT \cite{xu2022groupvit} and MixReorg.
    (a) GroupViT obtains image segmentation implicitly from image-text pairs to achieve cross-modal semantic alignment.
    (b) MixReorg explicitly constructs the fine-grained patch-text pairs data from the image-text pairs for free by mixing the patches from different images and preserving the correspondence between patches and text.
    }
    \vspace{-1.5em}
    \label{fig:CLIP_MixReorg}
\end{figure}

\section{Introduction}
\label{sec:intro}

Image segmentation has important applications in scenarios such as virtual presence, virtual try-on, movie post-production, and autonomous driving.
Currently, state-of-the-art semantic segmentation methods \cite{internimage, ren2022beyond, cheng2021mask2former} benefit from a large number of densely annotated data. 
However, the assumption of this closed-world setting requires that all categories of objects that appear in the test set are included in the training set. 
This heavy dependence on annotations limits that they can only work well in closed-set settings.
However, considering ubiquitous new concepts in real-world scenarios, learning an open-world segmentation model is more practical, but it is also more challenging. The open-world segmentation model is required to segment all entities and objects class-agnostically and exhaustively during training and be highly generalizable for aligning pixels with new semantics during testing.

Early methods achieve open-world semantic segmentation through few-shot learning \cite{cen2021deep} or unsupervised clustering \cite{nakajima2019incremental}. The former actually still assumes that the training and testing classes are in the same latent feature space, while the latter cannot guarantee the consistency of segmentation semantics. Recently, GroupViT \cite{xu2022groupvit} achieves state-of-the-art open-world segmentation performance using only text supervision.
It realizes the automatic grouping of image patches by vision-language contrastive learning (Figure \ref{fig:CLIP_MixReorg} (a)).
ViL-Seg \cite{liu2022open} implements image segmentation by introducing additional online clustering of visual embeddings for vision-language contrast.
Massive image-text pairs provide rich visual and textual semantics for open-world scenarios.
Similar to other CLIP-based \cite{CLIP} vision-language pre-training models (VLM) \cite{strudel2022weakly,yao2021filip,ICLIP}, although these methods achieve local information alignment of different modalities to a certain extent, they are still a computationally-based implicit matching strategy (fine-grained matching is learned by computing patch-text \cite{strudel2022weakly} or token-wise \cite{yao2021filip} similarity matrices).
Therefore, \textit{how to learn more fine-grained semantic alignment from image-text pair data becomes a key challenge for text-based supervised open-world segmentation tasks.}

\begin{figure}[t]
    \centering
    \includegraphics[width=1\linewidth]{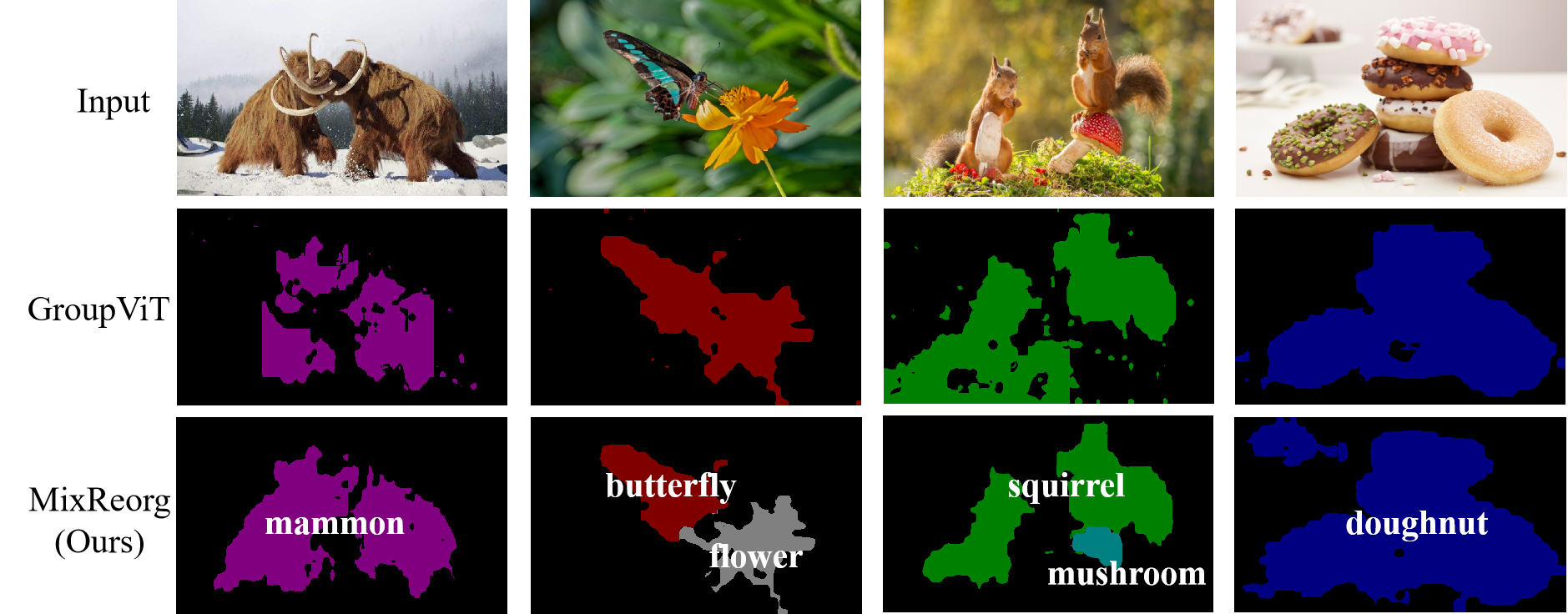}
    \caption{Visual comparison between MixReorg and GroupViT \cite{xu2022groupvit} on  images from the network. Our method can better handle open-world classes for segmentation task.}
    \vspace{-1.5em}
    \label{fig:open_world_visual}
\end{figure}

Inspired by the related work of mixed image modeling \cite{noroozi2016jigsaw, liu2022mixmim}, we propose a simple and novel cross-modal mixed image reconstruction mask learner. 
Specifically, as shown in Figure \ref{fig:CLIP_MixReorg} (b), MixReorg mixes patches from different images to generate mixed images.
Unlike previous methods for jigsaw puzzle \cite{noroozi2016jigsaw} or mixed image reconstruction \cite{liu2022mixmim} in the single visual modality, MixReorg's mixed patch reorganization is a cross-modal mask learner designed for semantic segmentation.
MixReorg preserves the correspondence between each patch and text when mixing image patches (the legend of Figure \ref{fig:CLIP_MixReorg} (b)).
In this way, we can obtain fine-grained patch-text pairs from the image-text data for free. 

However, there are still two challenges: \textit{(i)} the mixed image segmentation is easily disturbed by low-level features, which makes the model unable to realize patch reorganization of mixed images by high-level semantics; \textit{(ii})  each patch in mixed images are easily interfered by irrelevant patches from different images in the transformer layers, which may cause the image semantics to be difficult to match with the corresponding text.

For the first challenge, we propose two strategies of contextual mixing and progressive mixing to solve this problem.
The contextual mixing strategy allows each patch in the mixed image to obtain the global semantics of its original image in advance by adding a transformer layer before the mixing operation, thereby forcing the model to learn the mixed image reorganization from high-level semantics.
Furthermore, to further enhance the global information in the mixed image features, we propose to use the original image features to enhance the global semantics in the mixed image features.
For the second challenge, we present a mixing restoration strategy. It guarantees the semantic association of each patch token in the mixed image with the text through contrastive learning between the image recovered from mixed image and the text. In this way, the mutual interference between patches from different images in the mixed image can be effectively suppressed.

In general, MixReorg constructs a set of fine-grained patch-text pairs for free from image-text pair data, and successfully builds a cross-modal mixed image patch reorganization mask learner for open-world segmentation tasks.
The proposed MixReorg as a good mask learner also shows strong performance compared with the popular zero-shot semantic segmentation baselines,
achieving the performance of 50.5\%, 25.4\%, 23.6\% and 10.1\% mIoU on multi-scale evaluations on PASCAL VOC2012, PASCAL Context, MS COCO and ADE20K, respectively. 
The visualization in Figure \ref{fig:open_world_visual} shows that MixReorg significantly outperforms GroupViT \cite{xu2022groupvit} on open-world segmentation.
Our contributions can be summarized as follows:
\begin{itemize}
\vspace{-0.5em}
    \item We propose a novel and simple method that can easily construct patch-text data with fine-grained matching relationships from image-text data, thereby providing densely supervised information for open-world segmentation.
    \item For the constructed patch-text data, we propose a cross-modal mixed patch reorganization method. It successfully addresses the challenge of model failure due to mixed image segmentation susceptible to low-level features and irrelevant patches.
    \item The proposed MixReorg exhibits strong open-world segmentation performance and significantly outperforms current state-of-the-art zero-shot segmentation baselines.
\end{itemize}

\section{Related Work}

\noindent \textbf{VLM and Segmentation.} Recently, vision-language pre-training models \cite{CLIP, li2021align} have achieved great success. The models \cite{dong2022maskclip,CLIP,xu2022groupvit,yao2022detclip} trained with VLM are flexible and versatile, and can adapt to visual \cite{xu2022groupvit,mu2021slip, CLIP, strudel2022weakly} and multi-modal \cite{wang2021distilled, kim2021vilt, zhang2022glipv2} upstream and downstream tasks only by using the matching relationship of image-text data. 
This success has also been found in segmentation and has attracted the attention of lots of researchers \cite{xu2022groupvit, strudel2022weakly,xu2021simple,zhou2021denseclip}.
Because traditional semantic segmentation is limited by expensive manual dense annotation, VLMs are expected to break this limitation. Although the above methods achieve promising performance, using image-text sample-level matching relations to learn segmentation masks still faces the challenge of lacking local dense supervision information. 
In addition, there are some works \cite{yao2021filip, strudel2022weakly,ICLIP} that explore the alignment of multi-modal local information, but they are still computationally-dominated pseudo-local information correspondence without hard fine-grained supervision from the data level. Therefore, how to obtain finer-grained local supervision information from image-text data through data-level improvement is in great demand for semantic segmentation.

\begin{figure*}[t]
	\centering
	\includegraphics[width=0.9 \linewidth]{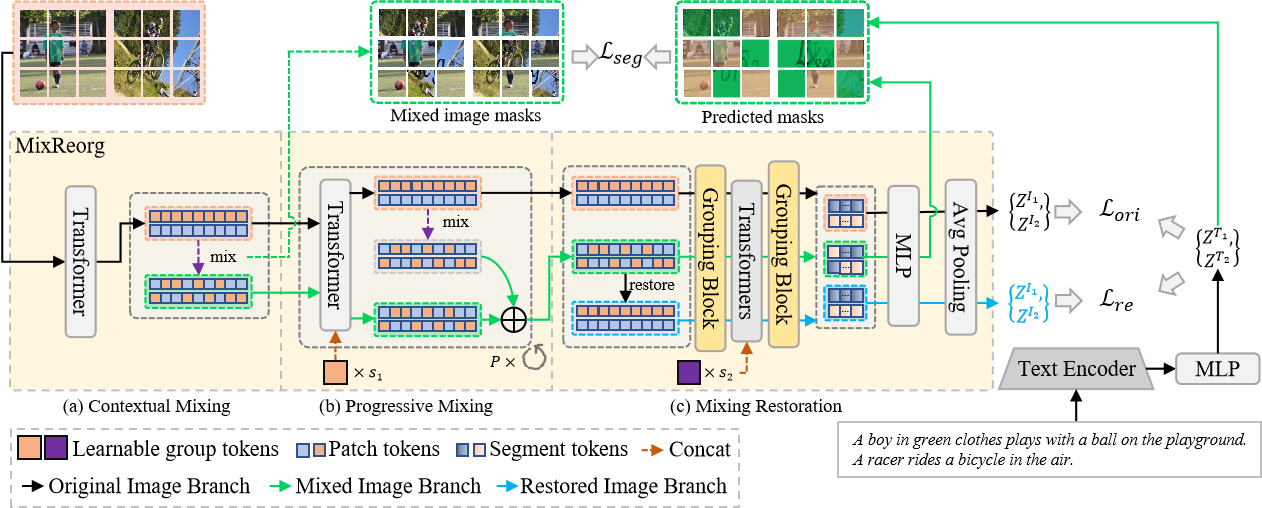}
	\vspace{-0.5em}
	\caption{The training pipeline and framework of MixReorg (take two images as an example).
    MixReorg's image encoder can be divided into three stages: (a) contextual mixing stage: a set of additional patch-text pairs with known segmentation mask is obtained by randomly mixing contextual patches from different images; (b) progressive mixing stage: 
    the original image features are used to enhance the global information of the mixed image features after mixing; 
    (c) mixing restoration stage: the original features, mixed features, and restored features are segmented through a two-stage grouping block \cite{xu2022groupvit}, and the corresponding segment tokens are obtained. 
    Note that we omit group tokens in the forward process for simplicity.
    During testing, MixReorg only needs to execute the original image branch.
    }
	\label{fig:OverallFramework}
	\vspace{-1.7em}
\end{figure*}
\noindent \textbf{Self-Supervision Strategies.} Self-supervision is an effective way to avoid the limitation of expensive manual annotations. It builds a self-supervised pipeline by fully mining the properties of the data itself. For example, self-supervision strategies such as masked image reconstruction \cite{he2022MAE, liu2022mixmim, li2022UM-MAE}, jigsaw puzzles \cite{noroozi2016jigsaw}, multi-view contrast \cite{DINO, zhou2021ibot, chen2020simpleclr} and angle recognition \cite{gidaris2018unsupervisedrotations} are widely adopted in the vision domain. 
Similar self-supervision strategies have been widely and successfully applied in natural language processing \cite{devlin2018bert, wang2022beit3}.
But the above methods are all designed for the single modality.
In contrast, semantic segmentation not only needs to consider the representation and segmentation of image features but also needs to consider cross-modal semantic alignment.
Therefore, how to draw more supervised information from image-text data by borrowing self-supervision strategies from the vision domain is very beneficial for semantic segmentation. Especially for the extraction of cross-modal fine-grained supervision information.

\noindent \textbf{Open-World Segmentation.} The open-world problem has been studied in the context of recognition \cite{bendale2015towards, liu2019large}, namely how to get a model trained only on a given closed-world dataset to also recognize new classes of objects. Similar settings are also used in object detection \cite{wang2021unidentified, joseph2021towards} and segmentation \cite{cen2021deep, nakajima2019incremental, xu2022groupvit}. For example, \cite{nakajima2019incremental} proposes an unsupervised open-world semantic segmentation; however, it obtains the mask of the image by a clustering method without any network parameter update.
On the other hand, VLMs \cite{CLIP,ICLIP,dong2022maskclip} exhibit strong performance and generalization ability with the help of massive image-text pair data.
Inspired by this, TSEG \cite{strudel2022weakly} and ICILP \cite{ICLIP} attempt to obtain fine-grained semantic alignment from image-text pairs to achieve image segmentation.
Similarly, GroupViT \cite{xu2022groupvit} introduces a set of learnable group tokens for ViT to group  patches and uses the generated segment tokens to align with text embeddings.
The massively available image-text pair data provides rich visual and textual semantics for open-world scenarios \cite{yao2022detclip}.
Therefore, open-world semantic segmentation based on text supervision can achieve more refined segmentation results at a lower cost of annotation. 
Based on the above observations, this paper follows the open-world semantic segmentation setting based on text supervision to further improve the performance of semantic segmentation.

\section{Methodology}
The overall framework of the MixReorg mask learner is shown in Figure \ref{fig:OverallFramework}. MixReorg is CLIP-based \cite{CLIP} and mainly consists of an image encoder and a text encoder. We use the text encoder from CLIP \cite{CLIP}. MixReorg's image encoder mainly has three stages: contextual mixing, progressive mixing, and mixing restoration (Sec. \ref{sec:overall_framework}).
Then, the loss composition of MixReorg is described in detail in Sec. \ref{sec:cross_modal_mixReorg}. Finally, the total loss is introduced in Sec. \ref{sec:all_loss}.

\subsection{MixReorg}
\label{sec:overall_framework}
\vspace{-0.3em}

\noindent\textbf{Contextual Mixing.}
As shown in Figure \ref{fig:OverallFramework}(a), in the contextual image patch mixing stage, patches from different images are randomly mixed to construct a set of mixed images with known segmentation masks.
According to the original image-text pairs, the patch-text correspondence of the mixed images is preserved, and the mixed image masks are used as the semantic segmentation labels of the mixed images. Similar to \cite{liu2022mixmim}, the mixed images only have randomly mixed patch tokens from different images at their same locations.
Specially, unlike \cite{liu2022mixmim}, MixReorg adopts a contextual information image patch mixing strategy.
We add a transformer layer before the image patch mixing operation to provide each patch the global image semantic which is closer to the text to create the coherence between patch and text. Meanwhile, it can preliminarily force the model to learn the mixed image patch reorganization from high-level features, thus effectively avoiding the interference of low-level features with the semantic learning of the model. 

Specifically, given a batch of image-text pairs $\{(x_i^I, x_i^T)\}_{i = 1}^B$. 
Following the design in ViT \cite{dosovitskiy2020vit}, we first split each input image into $N$ non-overlapping patches and linearly project each patch into a latent space. These projected patches are denoted as $\{\mathrm{p}_i\}_{i=1}^N$. 
For $M$ image-text pairs $\{(x_i^I, x_i^T)\}_{i = 1}^M$, MixReorg randomly mixes the patches from $M$ different images to construct $M$ mixed images, and the corresponding patch composition of $M$ mixed images can be denoted as
\vspace{-0.5em}
\begin{equation}
    \text{mix}(\{\{\mathrm{p}_i\}^N_{i=1}\}_{m=1}^M) = \{\{\mathrm{p}_{m,i}^j\}_{i=1}^N\}_{m=1}^M,1\leq j\leq M,
\end{equation}
where $\mathrm{p}_{m,i}^j$ denotes that the $i$-th patch of the $m$-th mixed image comes from the $j$-th image. Correspondingly, we keep the correspondence between each image patch and the text corresponding to the original image, resulting in a semantic segmentation dataset with patch-text correspondence.
The patch-text correspondence of the $m$-th mixed image can be expressed as $\{\mathrm{p}_{m,i}^j,x^T_j\}_{i=1}^N$.
Therefore, we obtain a set of sample-level image-text pairs $\{(x_i^I, x_i^T)\}_{i = 1}^B$ and a set of patch-level patch-text pairs $\{\{\mathrm{p}_{m,i}^j,x^T_j\}_{i=1}^N\}_{m=1}^M$.

\noindent \textbf{Progressive Mixing.} Mixed patches’ features
cannot improve after Contextual Mixing because of
the semantic mixing in mixed images. Since more layers in Contextual Mixing will lead to more parameters, we propose Progressive Mixing to enhance mixed features without additional parameters. As shown in Figure \ref{fig:OverallFramework}(b), in the progressive mixing phase, the patch tokens of normal and mixed images are concatenated with $s_1$ learnable group tokens $\{g_i\}_{i=1}^{s_1}$ respectively and fed to the multi-layer transformers independently. At the same time, the original features are used to enhance the contextual information of the mixed features.
The above process of an original image going through the $l$-th transformer layer can be represented as
\vspace{-0.5em}
\begin{equation}
    \{\{\hat{g}_i\}_{i=1}^{s_1},\{\mathrm{\hat{p}}_i\}_{i=1}^N\}=\text{Trans}_l([\{g_i\}_{i=1}^{s_1};\{\mathrm{p}_i\}_{i=1}^N]),
    \vspace{-0.5em}
\end{equation}
where $[\enspace;\enspace]$ denotes the concatenation operator.
Similarly, the output of the $m$-th mixed image through the $l$-th transformer layer can be expressed as
\vspace{-0.5em}
\begin{equation}
\begin{split}
    \{\{\hat{g}_i\}_{i=1}^{s_1},\{\mathrm{\hat{p}}_{m,i}^j\}_{i=1}^N\}=&\text{Trans}_{l}([\{g_i\}_{i=1}^{s_1};\{\mathrm{p}_{m,i}^j\}_{i=1}^N+\\&\text{mix}(\{\{\{\mathrm{p}_i\}^N_{i=1}\}_{m=1}^M\}_{l-1})_m]). 
    \vspace{-1.5em}
\end{split}
\end{equation}
\noindent \textbf{Mixing Restoration.}
Although we can achieve patch-text alignment by mixed images, there are a lot of different semantics
in one mixed image which will interfere with each other.
Therefore, this requires the patches from different images in a mixed image still need to maintain a semantic match with the corresponding text.

To this end, as shown in Figure \ref{fig:OverallFramework}(c), in the mixing restoration phase, MixReorg also restores the mixed image according to the patch position of the image before mixing. The original features $\{\mathrm{p}_i\}_{i=1}^N$, mixed features $\{\mathrm{p}_{m,i}^j\}_{i=1}^N$, and restored features $re(\{\mathrm{p}_{m,i}^j\}_{i=1}^N)$ are segmented through a two-stage grouping block \cite{xu2022groupvit}, and the corresponding segment tokens $\{\text{seg}_i\}_{i=1}^{s_2}$ are obtained. These segment tokens are fed into multiple transformer layers and then projected to the same dimensionality $D$ as text embeddings $Z^T \in \mathbb{R}^{B\times D}$ through an MLP.
\subsection{Cross-Modal Mixed Patch Reorganization Loss}
\label{sec:cross_modal_mixReorg}
\vspace{-0.3em}

\begin{figure}[t]
    \centering
    \includegraphics[width=1 \linewidth]{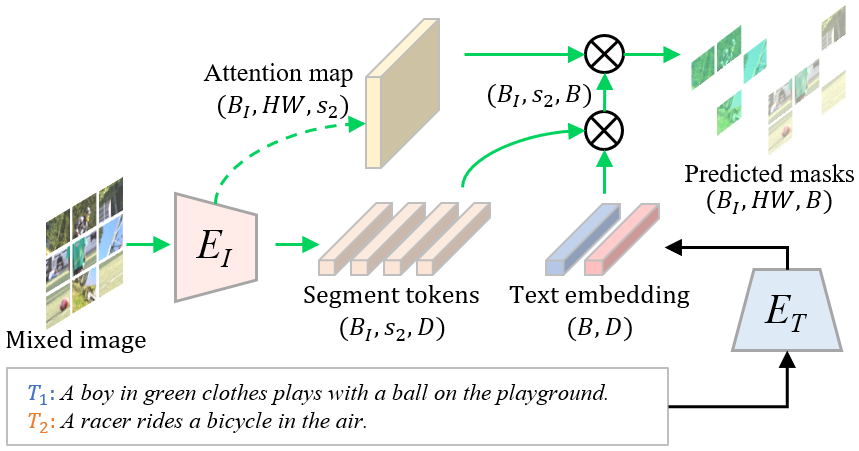}
    \caption{
    Cross-modal mixed patch reorganization, which combines attention maps and segmentation tokens from the image encoder, and text embeddings to reorganize and predict segmentation masks for mixed images. Where $B_I=M$ means that every $B_I$ images are mixed.
    For simplicity, we take a mixed image generated by mixing two images as an example.
    }
    \vspace{-1.5em}
    \label{fig:predicted_masks}
\end{figure}

\noindent\textbf{Mixed Segmentation Loss.}
Cross-modal mixed patch reorganization is a core module designed in MixReorg for semantic segmentation. It provides the model with more refined local alignment information by using a constructed mixed image dataset with patch-text correspondence. We expect the model to learn more fine-grained semantic alignment with the help of the constructed patch-text pairs. And the calculation process of the mixed image segmentation mask is shown in Figure \ref{fig:predicted_masks}, where every $B_I$ (\textit{e.g}., $B_I=M$) images are mixed. The mixed image segment tokens and the text embeddings of the whole batch are used to compute the similarity $S\in\mathbb{R}^{B_I\times s_2\times B}$ for the text, where $\otimes$ denotes matrix multiplication. Since we adopt a two-stage grouping block \cite{xu2022groupvit}, we have the attention map $A\in\mathbb{R}^{B_I\times HW\times s_2}~~(HW=N)$, which contains the grouping relationship between $HW$ patches and $s_2$ segment tokens. Further, by $A\otimes S$, we can predict the segmentation mask $M_p\in \mathbb{R}^{B_I\times HW \times B}$ of the mixed image. 
Finally, we compute the cross-entropy loss between the mixed image masks $M_{mix}$ and the prediction masks $M_p$, formulated as
\vspace{-0.5em}
\vspace*{-0.5\baselineskip}
\begin{equation}
    \mathcal{L}_{seg}=\mathcal{L}_{CE}(M_p, M_{mix}),
    \vspace{-0.5em}
\end{equation}
where $\mathcal{L}_{CE}(\boldsymbol{p},\boldsymbol{q})=-\sum_i \boldsymbol{q}_i \text{log}(\boldsymbol{p}_i)$ is the cross-entropy loss of output $\boldsymbol{p}$ and target $\boldsymbol{q}$.

\noindent\textbf{Restoration Contrastive Loss.}
Furthermore, we consider cross-modal semantic contrastive learning of mixing restoration features and text embeddings.
Specifically, in order to take full advantage of the correspondence provided by the image-text pair and enhance the model's ability to align semantically across modalities, MixReorg computes the contrastive loss and the multi-label image-text contrastive loss \cite{xu2022groupvit} between the output of the restored features branch and the text embeddings, respectively.
The calculation of the total image-text contrastive loss is as follows
\vspace{-0.5em}
\begin{equation}
    \mathcal{L}_{I\leftrightarrow T}^{re} = \mathcal{L}_{I\rightarrow T}^{re} +\mathcal{L}_{I\leftarrow T}^{re},
    \vspace{-0.5em}
\end{equation}
where the image-to-text contrastive loss is defined as
\vspace{-0.5em}
\begin{equation}
    \mathcal{L}_{I\rightarrow T}^{re} = -\frac{1}{B} \sum_{i = 1}^{B} \log \frac{\exp \left(z_{i}^{I} \cdot z_{i}^{T} / \tau\right)}{\sum_{j = 1}^{B} \exp \left(z_{i}^{I} \cdot z_{j}^{T} / \tau\right)},
    \vspace{-0.5em}
\end{equation}
and the text-to-image contrastive loss is defined as
\vspace{-0.5em}
\begin{equation}
    \mathcal{L}_{I\leftarrow T}^{re}=-\frac{1}{B} \sum_{i=1}^{B} \log \frac{\exp \left(z_{i}^{T} \cdot z_{i}^{I} / \tau\right)}{\sum_{j=1}^{B} \exp \left(z_{i}^{T} \cdot z_{j}^{I} / \tau\right)},
    \vspace{-0.5em}
\end{equation}
where $\tau$ is a learnable temperature parameter, and $z_{i}^{I}$ and $z_{i}^{T}$ are image and text embedding for image-text pairs $\{(x_i^I, x_i^T)\}_{i = 1}^B$.
In addition, we calculate the multi-label contrastive loss of the restored features branch as follows
\vspace{-0.5em}
\begin{equation}
    \mathcal{L}_{I\leftrightarrow \{T_k\}_{k=1}^K}^{re} = \mathcal{L}_{I\rightarrow \{T_k\}_{k=1}^K}^{re} +\mathcal{L}_{I\leftarrow \{T_k\}_{k=1}^K}^{re},
    \vspace{-0.5em}
\end{equation}

where $\{T_k\}_{k=1}^K$ is an additional $K$ text labels generated using the “prompting engineering" mechanism in \cite{CLIP}, 
\vspace{-0.5em}
\begin{equation}
    \mathcal{L}_{I\rightarrow \{T_k\}_{k=1}^K}^{re}=-\frac{1}{B} \sum_{i=1}^{B} \log \frac{\sum_{k=1}^{K} \exp \left(z_{i}^{I} \cdot z_{i}^{T_{k}} / \tau\right)}{\sum_{k=1}^{K} \sum_{j=1}^{B} \exp \left(z_{i}^{I} \cdot z_{j}^{T_{k}} / \tau\right)}
    \vspace{-0.5em}
\end{equation}
and 
\vspace{-0.5em}
\begin{equation}
    \mathcal{L}_{I\leftarrow \{T_k\}_{k=1}^K}^{re}=-\frac{1}{K B} \sum_{k=1}^{K} \sum_{i=1}^{B} \log \frac{\exp \left(z_{i}^{T_{k}} \cdot z_{i}^{I} / \tau\right)}{\sum_{j=1}^{B} \exp \left(z_{i}^{T_{k}} \cdot z_{j}^{I} / \tau\right)}.
    \vspace{-0.5em}
\end{equation}
The total contrastive loss of the restored features and the text embeddings is as follows
\vspace{-0.5em}
\begin{equation}
    \mathcal{L}_{re} = \mathcal{L}_{I\leftrightarrow T}^{re} +\mathcal{L}_{I\leftrightarrow \{T_k\}_{k=1}^K}^{re}.
    \vspace{-0.5em}
    \label{eq:loss_restoration}
\end{equation}
In summary, the total cross-modal mixed image patch reorganization loss is as follows
\vspace{-0.5em}
\begin{equation}
    \mathcal{L}_{mixed}=\mathcal{L}_{seg}+\mathcal{L}_{re}.
    \vspace{-1em}
\end{equation}

\begin{table}[t]
    \centering
    \begin{tabular}{l|c|c}
    \hline
    \multirow{2}{*}{Arch.} & \multirow{2}{*}{Method} & Mask  \\
    &&mIoU (\%) \\\hline
    \multirow{4}{*}{ViT \cite{dosovitskiy2020vit}$^*$}& pixel-wise& 20.1 \\
      & K-means& 25.0 \\
      & Mean-shift& 20.7\\
      & Spectral clustering & 19.7 \\
    \hline
    
    GroupViT \cite{xu2022groupvit} & -  & 41.1\\
    ViewCo \cite{ren2023viewco} & - & 45.7\\
    MixReorg (ours) & - & \textbf{47.9}\\
    \hline
    \end{tabular}
    \vspace{-0.5em}
    \caption{Comparison with zero-shot semantic segmentation baselines on PASCAL VOC. GroupViT\cite{xu2022groupvit} and MixReorg are trained on CC12M. The superscript $^*$ means the results are from \cite{xu2022groupvit}.
    \vspace{-1.5em}
    }
    \label{tab:zero-shot}
\end{table}
\subsection{Overall Loss Function}
\label{sec:all_loss}
\vspace{-0.3em}
Similar to Eq. (\ref{eq:loss_restoration}), the total contrastive loss between the original image features and the text embeddings is as follows
\begin{equation}
    \mathcal{L}_{ori} = \mathcal{L}_{I\leftrightarrow T}^{ori} +\mathcal{L}_{I\leftrightarrow \{T_k\}_{k=1}^K}^{ori}.
    \vspace{-0.5em}
\end{equation}
Finally,
the total loss of MixReorg is 
\vspace{-0.5em}
\begin{equation}
    \mathcal{L}= \mathcal{L}_{mixed}+\mathcal{L}_{ori}.
    \vspace{-1em}
\end{equation}
When testing, MixReorg only needs to execute the original image branch (the solid black line in Figure \ref{fig:OverallFramework}), so it does not add any extra testing time.

\begin{table*}[!t]
    \centering
    \begin{tabular}{l|lcc|c|c|c} 
    \toprule
    & \multicolumn{3}{c|}{ Pre-training } & & \multicolumn{2}{c}{Transfer (mIoU (\%))} \\\hline
    Arch. & Model & Dataset & Supervision & Zero-Shot & \makecell{PASCAL\\VOC} & \makecell{PASCAL\\Context}  \\\hline 
    \multirow{6}{*}{ViT} & DeiT \cite{deit} & ImageNet & class & \XSolidBrush & $53.0$ & $35.9$  \\\cline{2-7}
    & DINO \cite{DINO} & ImageNet & self & \XSolidBrush & $39.1$ & $20.4$ \\
    & DINO & CC12M+YFCC & self & \XSolidBrush & $37.6$ & $22.8$ \\
    & MoCo \cite{moco} & ImageNet & self & \XSolidBrush & $34.3$ & $21.3$  \\
    & MoCo & CC12M+YFCC & self & \XSolidBrush & $36.1$ & $23.0$  \\ \hline 
    \multirow{9}{*}{CLIP} & SLIP$^*$ \cite{mu2021slip} & LAION-20M & text \& self & \Checkmark & - & $12.3$  \\
    & CLIP-MAE$^*$ \cite{dong2022maskclip} & LAION-20M & text \& self & \Checkmark & - & $16.8$  \\
    & MaskCLIP \cite{dong2022maskclip} & LAION-20M & text \& self & \Checkmark & - & $17.7$ \\
    & ViewCo \cite{ren2023viewco} & CC12M & text \& self & \Checkmark & 45.7 & 20.8
    \\\cline{2-7}
    & MaskCLIP \cite{zhou2021denseclip} & CLIP-400M & text & \Checkmark  & - & $21.7$\\
    & CLIP$^*$ \cite{CLIP}  & LAION-20M & text & \Checkmark  & - & $13.5$  \\
        & GroupViT\cite{xu2022groupvit}  & CC12M+YFCC & text & \Checkmark  & \textbf{51.2} & $22.3$  \\\cline{2-7}
    & GroupViT  & CC12M & text & \Checkmark & \makecell{41.1\\$45.5^{\dagger}$} & \makecell{18.2\\$19.2^{\dagger}$} \\\cline{2-7}
    & \multirow{2}{*}{MixReorg (ours)}   & \multirow{2}{*}{CC12M} & \multirow{2}{*}{text} & \multirow{2}{*}{\Checkmark} & \textbf{47.9} $(6.8\uparrow)$  & \textbf{23.9} $(5.7\uparrow)$  \\
    &&&&&\textbf{50.5}$^{\dagger} (5.0\uparrow)$ & \textbf{25.4}$^{\dagger} (6.2\uparrow)$  \\
    \bottomrule
    
    \end{tabular}
    \vspace{-0.5em}
    \caption{Performance comparison on PASCAL VOC \cite{everingham2010pascal} and PASCAL Context \cite{mottaghi2014role}.
    Zero-shot means that the model is directly transferred to the semantic segmentation task without any fine-tuning on the target dataset. The superscript $^*$ denote the results are from \cite{dong2022maskclip}. $^{\dagger}$ indicates the results of the multi-scale evaluation. 
    }
    \vspace{-2em}
    \label{tab:SoTA}
\end{table*}

\begin{table}[!t]
    \centering
    \begin{tabular}{l|c|c}
    \hline
    \multirow{2}{*}{Model} & \multirow{2}{*}{\makecell{Pre-training\\Dataset}} & Transfer   \\
    &&mIoU (\%) \\\hline
    ViewCo \cite{ren2023viewco} & CC12M & 20.6 \\
    GroupViT \cite{xu2022groupvit} & CC12M+YFCC & 20.9\\\hline
    GroupViT                       & CC12M & \makecell{18.4\\21.1$^\dagger$}\\\hline
    MixReorg (ours) & CC12M &\makecell{\textbf{21.3} ($2.9\uparrow$)\\ \textbf{23.6}$^\dagger$ ($2.5\uparrow$)}\\
    \hline
    \end{tabular}
    \vspace{-0.5em}
    \caption{Performance comparison on COCO \cite{lin2014microsoft}. $^{\dagger}$ indicates the results of the multi-scale evaluation.
    }
    \label{tab:coco}
    \vspace{-2em}
\end{table}

\section{Experiments}
\subsection{Implementation Details}
\vspace{-0.5em}

\label{sec:detail}
\noindent\textbf{Architecture.} The image encoder of MixReorg is based on a 2-stage GroupViT \cite{xu2022groupvit} with 12 transformer layers, while adding one transformer layer before the mix operation. The size of the input image is $224 \times 224$, the patch size is $16 \times 16$ and the hidden dimensionality is $384$. The model outputs 32 segment tokens (\textit{i.e.}, $s_2=32$ and $s_1=64$). Following \cite{CLIP, xu2022groupvit}, the text encoder of MixReorg consists of 12 layers of transformer with the hidden feature dimensionality of 256.

\noindent\textbf{Training and Inference.} During the training phase, we use CC12M \cite{changpinyo2021cc12m} as the training dataset, which contains 12M image-text pairs. We apply the mix operation for every 16 images (\textit{i.e.}, $M=16$). Following \cite{xu2022groupvit, CLIP}, our batch size is 4096. We set the weight for each loss function to 1. During the inference phase, only the original image branch is executed. More details can be seen in Appendix.

\noindent\textbf{Open-World Semantic Segmentation.} 
We evaluate the performance of MixReorg on the open-world segmentation task on four commonly used open semantic segmentation datasets PASCAL VOC 2012 \cite{everingham2010pascal}, PASCAL Context \cite{mottaghi2014role}, COCO \cite{lin2014microsoft}, and ADE20K \cite{zhou2017scene}.
They contain 20, 59, 80, and 150 foreground classes, respectively, with validation images of 1.5K, 5K, 5K, and 2K, respectively.
MixReorg is transferred to the target dataset in a zero-shot manner without any fine-tuning. Following GroupViT \cite{xu2022groupvit}, MixReorg obtains the corresponding segmentation of the image through the learned group token.

\subsection{Comparisons with Existing Methods}
\vspace{-0.5em}
\label{sec:compare}
\noindent\textbf{Comparison with Zero-Shot Baselines.} In Table \ref{tab:zero-shot}, we present a comparison of four ViT-based baselines, which utilize the image-text contrastive loss defined in CLIP \cite{CLIP} to train the vision and text encoders. These baselines employ pixel-wise, k-means, mean-shift, and spectral clustering strategies, respectively. Additionally, we include GroupViT \cite{xu2022groupvit}, which employs a bottom-up grouping method. The results in Table \ref{tab:zero-shot} demonstrate that MixReorg outperforms both the ViT baselines and GroupViT (41.1\% \emph{vs.} 47.9\%), indicating that MixReorg is effective in enhancing the segmentation ability of the model.

\begin{table}[!t]
    \centering
    \begin{tabular}{l|c|c}
    \hline
    \multirow{2}{*}{Model} & \multirow{2}{*}{\makecell{Pre-training\\Dataset}} & Transfer   \\
    &&mIoU (\%) \\\hline
    ALIGN$^a$ \cite{align}& ALIGN-1800M& \textbf{9.7} \\
    ALIGN$^a$  & HQITP-134M& 7.5 \\
    CLIP$^a$ \cite{CLIP} & HQITP-134M& 5.1\\
    CLIP  &CLIP-400M & 5.8 \\
    CLIP$^b$  &LAION-20M & 7.7 \\
    SLIP$^b$ \cite{mu2021slip} & LAION-20M & 6.8 \\
    \hline
    GroupViT \cite{xu2022groupvit} & CC12M & \makecell{5.8\\6.7}\\\hline
    MixReorg (ours) & CC12M &\makecell{\textbf{8.7} ($2.9\uparrow$)\\ \textbf{10.1}$^\dagger$ ($3.4\uparrow$)}\\
    \hline
    \end{tabular}
    \vspace{-0.5em}
    \caption{Performance comparison on ADE20K \cite{zhou2017scene}. The superscript $^a$ and $^b$ denote the results are from  \cite{ranasinghe2022perceptual_CLIPpy} and \cite{dong2022maskclip}, respectively.
    }
    \label{tab:ade}
    \vspace{-2em}
\end{table}
\noindent\textbf{Comparison with SoTA Methods.}
We conduct a comprehensive evaluation of MixReorg against fully-supervised methods \cite{deit}, self-supervised methods \cite{DINO, moco}, vision-language contrastive learning baselines \cite{zhou2021denseclip, CLIP, xu2022groupvit}, and baselines combining vision-language contrastive learning with self-supervised learning \cite{mu2021slip, dong2022maskclip, ren2023viewco}. Table \ref{tab:SoTA} and Table \ref{tab:coco} summarizes the comparison results on PASCAL VOC, PASCAL Context, and COCO datasets, both in single-scale and multi-scale evaluation settings. The results show that MixReorg outperforms all the methods by a significant margin, demonstrating its effectiveness in semantic segmentation.
In comparison to GroupViT, MixReorg that pre-trains on CC12M yields substantial performance improvements of 6.8\% mIoU, 5.7\% mIoU, and 2.9\% mIoU for single-scale evaluation, and 5.0\% mIoU, 6.2\% mIoU, and 2.5\% mIoU for multi-scale evaluation on PASCAL VOC, PASCAL Context, and COCO, respectively. MixReorg also outperforms GroupViT which pre-trains on CC12M and YFCC with less data. Additionally, MixReorg has a clear advantage over the methods which rely on additional supervision information in the vision branch.
Furthermore, we evaluate the performance of MixReorg on the ADE20K dataset, as shown in Table \ref{tab:ade}. MixReorg outperforms GroupViT by a significant margin (8.7\% \emph{vs.} 5.8\%), demonstrating its superior performance in complex segmentation tasks.\\
\noindent\textbf{Image Classification.} We evaluate the zero-shot classification performance of MixReorg on ImageNet. As shown in Table \ref{tab:cls}, MixReorg significantly outperforms GroupViT, indicating that MixReorg  achieves better image-text alignment through fine-grained mask learning.
\begin{table}[ht]
    \vspace{-1em}
    \centering
    \begin{tabular}{l|cc}
    \hline
    \multirow{2}{*}{Arch.}  & \multicolumn{2}{c}{Zero-shot}  \\
    &Acc@1 (\%)& Acc@5 (\%)\\\hline
    GroupViT \cite{xu2022groupvit}   & 37.5 & 65.5\\
    MixReorg(ours)  & \textbf{38.8} & \textbf{66.7}\\
    \hline
    \end{tabular}
    \vspace{-0.5em}
    \caption{Zero-shot classification performance on ImageNet.
    }
    \label{tab:cls}
    \vspace{-1.5em}
\end{table}
\vspace{-0.5em}
\subsection{Ablation Study}
\vspace{-0.5em}
\label{sec:ablation}
\noindent\textbf{Contextual Mixing.} 
In Table \ref{tab:array}, we perform ablation on contextual mixing (CM) strategies.
Specifically, we first ablate the parameter variation of the model.
As the results show, by adding a transformer layer to GroupViT's image encoder (\textit{i.e.}, GroupViT+), the performance of the model remains  consistent (18.4\% \textit{vs.} 18.2\%).
This shows that simply increasing the number of parameters does not improve the performance of the model.
Further, we also compared MixReorg with only CM (\textit{i.e.}, row 3) and GroupViT+. They have the same amount of parameters, but MixReorg significantly outperforms GroupViT+ with the help of CM (19.3\% \textit{vs.} 18.2\%). This shows that it is beneficial to help the model obtain more global semantic information in the early stage of the model.
For MixReorg, it can be found that MixReorg without CM assistance (row 4) will degenerate to a similar performance to GroupViT, which indicates that CM plays a crucial role in MixReorg's mixed segmentation module (19.3\% \textit{vs.} 18.0\% \textit{vs.} 18.4\%).
This is mainly because CM can help MixReorg acquire global semantic information early in the model, thus forcing the model to learn mixed image reorganization from high-level semantic, which frees the model from low-level semantic information (\textit{e.g.}, texture and color, etc.) to obtain trivial solutions.


\begin{table}[!h]
\vspace{-0.5em}
    \centering
    \begin{tabular}{l|ccc|c}
    \hline
       Method & CM & $\mathcal{L}_{seg}$ &$\mathcal{L}_{re}$ & mIoU (\%) \\\hline
        GroupViT  & -        &-         &-           &18.4  \\
        GroupViT+  & -        &-         &-           &18.2  \\\hline
        \multirow{4}{*}{MixReorg} & \checkmark &     &            &19.3  \\ 
                                &           &\checkmark &            &18.0 \\
                                &\checkmark &\checkmark &            & 20.5 \\
                                &\checkmark   &           & \checkmark & 20.3 \\
                                &\checkmark&\checkmark & \checkmark & \textbf{21.3} \\
    \hline
    \end{tabular}
    \vspace{-0.5em}
    \caption{On COCO, MixReorg's ablation study on contextual mixing (CM) and the loss functions. GroupViT+ means adding one transformer layer at the 1-st stage of GroupViT. For MixReorg, without CM means images are mixed prior to passing through the transformer layer that we add.}
    \label{tab:array}
    \vspace{-1em}
\end{table}
\noindent\textbf{Ablation of Losses.} 
In Table \ref{tab:array}, we also conduct an ablation study on each loss that MixReorg uses. Specifically, $\mathcal{L}_{seg}$ improves the performance by 1.2\% mIoU (20.5\% \emph{vs.} 19.3\%). This means that $\mathcal{L}_{seg}$ plays a significant role. The model learns to distinguish different semantics in the image through $\mathcal{L}_{seg}$. Additionally, $\mathcal{L}_{re}$ improves the performance of the model (20.3\% \emph{vs.} 19.3\%). It helps the patches of the mixed image to maintain consistency between its original image semantics and corresponding text. Furthermore, by combining $\mathcal{L}_{seg}$ and $\mathcal{L}_{re}$ based on CM, the performance of MixReorg can be further improved (21.3\% \emph{vs.} 20.5\% \emph{vs.} 20.3\%), which illustrates that CM and two loss functions are strongly related. CM is fundamental to achieving patch-text alignment since it provides global information to each patch, while $\mathcal{L}_{seg}$ provides fine-grained semantic alignment ability and $\mathcal{L}_{re}$ assisting $\mathcal{L}_{seg}$ keeping the patches' original semantic, free from the interference from different images.


\begin{figure}[!t]
    \vspace{-0.5em}
    \centering
    \includegraphics[width=1\linewidth]{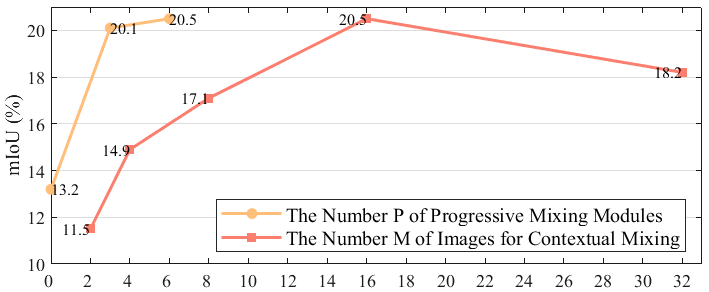}
    \vspace{-1.5em}
    \caption{On COCO, MixReorg's ablation study on the number of progressive mixings and the number of images for the contextual mixing operation. (a) Yellow line: Ablation study on the number $P$ of the progressive mixing modules. We replace one progressive mixing module with one transformer layer to maintain the model size. (b) Red line: Ablation study on the number $M$ of images for each contextual mixing operation.}
    \label{fig:tm_Ablation}
    \vspace{-2em}
\end{figure}

\begin{figure*}[t]
\vspace{-0.5em}
    \centering
    \includegraphics[width=0.9\linewidth]{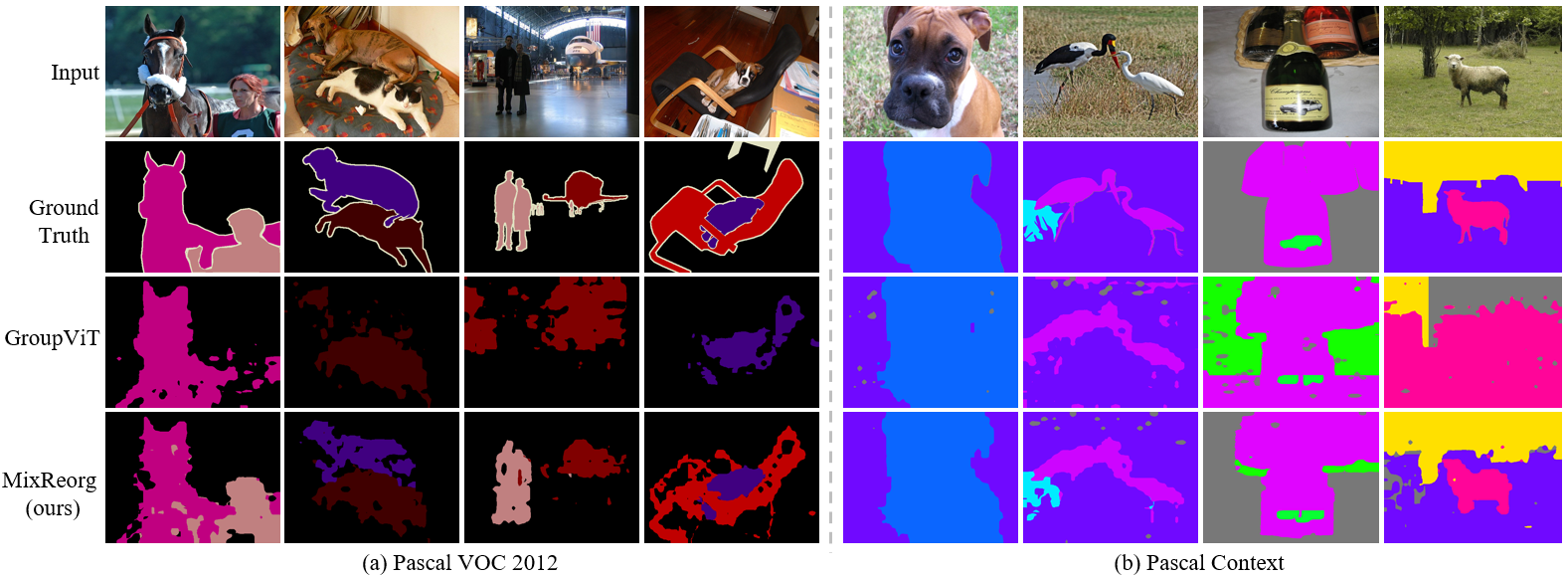}
    \vspace{-0.5em}
    \caption{Comparison of semantic segmentation results on PASCAL VOC 2012 and PASCAL Context.}
    \label{fig:voc_seg}
    \vspace{-2em}
\end{figure*}
\noindent\textbf{Number of Images for Mixing.} In Figure \ref{fig:tm_Ablation} (red line), we observe the performance impact of the number of images $M$ used for contextual mixing operation. It can be observed that $M=16$ is optimal. As $M$ increases,  mixed images contain more semantic categories, which is helpful for the model in learning semantic grouping (20.5\% \emph{vs.} 17.1\%). However, increasing $M$ beyond a certain threshold (\emph{e.g.}: $M = 32$) causes semantic representation in the mixed image to be insufficient due to resolution constraints, thereby interfering with model learning (20.5\% \emph{vs.} 18.2\%).

\noindent \textbf{Progressive Mixing Module.} In Figure \ref{fig:tm_Ablation} (yellow line), we study the number $P$ of the progressive mixing modules. We add one transformer layer when removing one progressive mixing module to maintain the model size. It can be seen that the model is optimal when the number $P$ of the progressive mixing modules is 6. The progressive mixing improves
over $P=0$ by about 7\% mIoU ($P = 3$ \emph{vs.} $P=0$). When $P=0$, the original image is not used to enhance the mixed image after the mixing operation. In this case, the lack of global information on the original image hinders the learning of the model. Obviously, with the increase of the progressive mixing modules, the semantics of the mixed image features become clearer, which is thus more conducive for model learning to distinguish different semantics, thus improving the model segmentation ability.

\vspace{-0.5em}
\subsection{Visualization}
\vspace{-0.5em}
\label{sec:vis}

\noindent\textbf{Qualitative Results.} In Figure \ref{fig:voc_seg}, we illustrate zero-shot semantic segmentation examples predicted by GroupViT and MixReorg to verify the segmentation capability of our method. As shown in Figure \ref{fig:voc_seg}(a), MixReorg can handle more complex segmentation examples which have different classes in one image, showing that our method can better perceive fine-grained semantics. In addition, as shown in Figure \ref{fig:voc_seg}(b), MixReorg's segmentation quality of stuff classes is significantly better than GroupViT.
In a word, MixReorg has a stronger ability of high-level semantic understanding and segmentation.

\begin{figure}[ht]
    \centering
    \subfloat[Image Reorganization]{\includegraphics[width=0.44\linewidth]{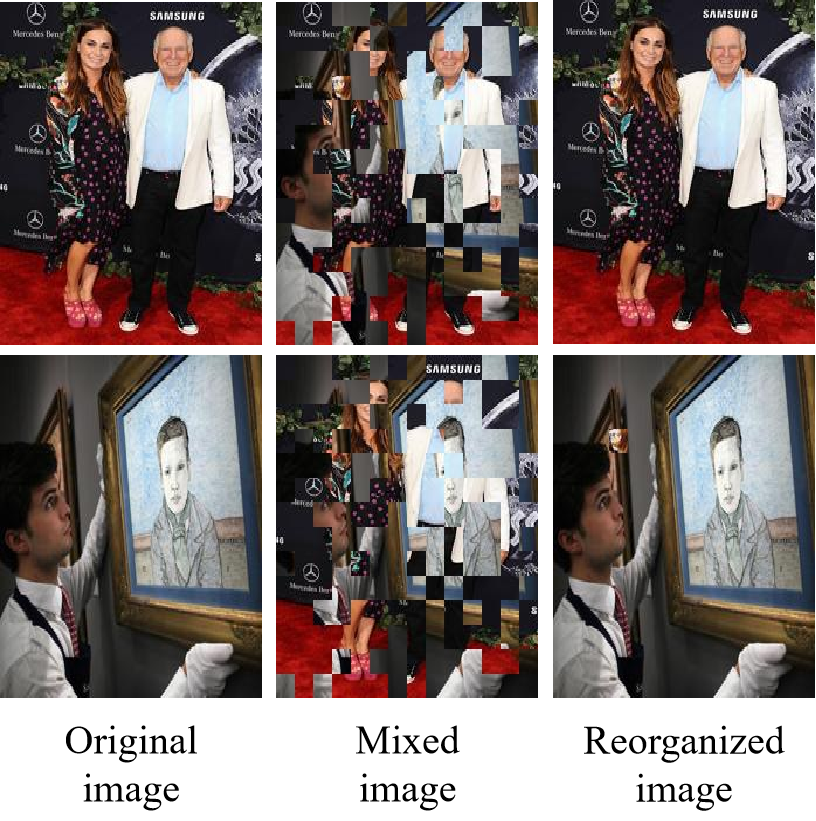}}
    \hspace{0.5em}
    \subfloat[Confusion Matrix] {\includegraphics[width=0.5\linewidth]{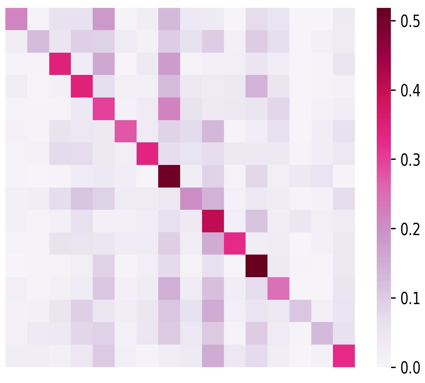} }
    \vspace{-1em}
    \caption{
    Mixed image reorganization and the confusion matrix. (a) We use the segmentation mask predicted by MixReorg on the mixed image to obtain the reorganized image. (b) Taking $M=16$ as an example, the confusion matrix $CM$ of the patch segmentation of the mixed images. $CM_{ij}$ represents the proportion of patches belonging to the $i$-th image in the mixed image that are classified into the $j$-th image category.
    }
    \label{fig:Mix_Restored}
    \vspace{-2em}
\end{figure}
\noindent\textbf{Mixed Patch Reorganization.} We visualize the reorganized images from mixed images according to the mask predictions from two mixed images in Figure \ref{fig:Mix_Restored}(a). It can be seen that except for a few patches, MixReorg can correctly segment most image patches into their corresponding original semantics. In Figure \ref{fig:Mix_Restored}(b), the confusion matrix of the prediction for one mixed image indicates that MixReorg can effectively align patches with text.
\vspace{-1em}
\section{Discussion}
\vspace{-0.5em}
\noindent\textbf{Conclusion.} We propose a patch-text data construction method with dense matching for image-text data and a cross-modal mixed image patch reorganization mask learner for mixed images to achieve fine-grained semantic alignment in open-world segmentation. MixReorg shows superior performance in open-world scenarios.

\noindent\textbf{Limitations.} There are two issues that we should explore to improve MixReorg. First, since we use contextual mixing to create additional dat, the computational budget is increased during the training phase. Second, although Mixreorg successfully constructs patch-text data for semantic segmentation, there is still a gap between it with pixel-level data.
\vspace{-1em}
\section{Acknowledgment}
\vspace{-1em}
This work was supported in part by National Key R\&D Program of China under Grant No. 2020AAA0109700,  Guangdong Outstanding Youth Fund (Grant No. 2021B1515020061), Shenzhen Science and Technology Program (Grant No. RCYX20200714114642083),
Shenzhen Fundamental Research Program(Grant No. JCYJ20190807154211365), Nansha Key RD Program under Grant No.2022ZD014 and Sun Yat-sen University under Grant No. 22lgqb38 and 76160-12220011, CAAI-Huawei MindSpore Open Fund. We thank MindSpore for the partial support of this work, which is a new deep learning computing framwork\footnote{https://www.mindspore.cn/}.
\clearpage
{\small
\bibliographystyle{ieee_fullname}
\bibliography{egbib}
}

\clearpage

\end{document}